\documentclass[runningheads]{llncs}
\usepackage[T1]{fontenc}
\usepackage{graphicx,verbatim}
\usepackage{amsmath}
\usepackage{hhline}
\usepackage{hyperref} 
\begin{document}
\titlerunning{WeakSupCon}
\title{WeakSupCon: Weakly Supervised Contrastive Learning for Encoder Pre-training}
%
\author{Bodong Zhang\inst{1,2}\thanks{Corresponding author. Email: bodong.zhang@utah.edu}\orcidID{0000-0001-9815-0303} \and
Hamid Manoochehri\inst{1,2}\orcidID{0009-0005-0478-7925} \and
Xiwen Li\inst{2}\orcidID{0009-0005-6139-0668} \and
Beatrice S. Knudsen\inst{3}\orcidID{0000-0002-7589-7591} \and
Tolga Tasdizen\inst{1,2}\orcidID{0000-0001-6574-0366}}
\authorrunning{B. Zhang et al.}
%
\institute{Department of Electrical and Computer Engineering, University of Utah, Salt Lake City, UT, USA \and
Scientific Computing and Imaging Institute, University of Utah, SLC, UT, USA \and
Department of Pathology, University of Utah, SLC, UT, USA}

\maketitle              
\begin{abstract}
Weakly supervised multiple instance learning (MIL) is a challenging task given that only bag-level labels are provided, while each bag typically contains multiple instances. This topic has been extensively studied in histopathological image analysis, where labels are usually available only at the whole slide image (WSI) level, while each WSI could be divided into thousands of small image patches for training. The dominant MIL approaches focus on feature aggregation and take fixed patch features as inputs. However, weakly supervised feature representation learning in MIL settings is always neglected. Those features used to be generated by self-supervised learning methods that do not utilize weak labels, or by foundation encoders pre-trained on other large datasets. In this paper, we propose a novel weakly supervised feature representation learning method called Weakly Supervised Contrastive Learning (WeakSupCon) that utilizes bag-level labels. In our method, we employ multi-task learning and define distinct contrastive losses for samples with different bag labels. Our experiments demonstrate that the features generated using WeakSupCon with limited computing resources significantly enhance MIL classification performance compared to self-supervised approaches across three datasets. Our WeakSupCon code is available at \url{github.com/BzhangURU/Paper_WeakSupCon}

\keywords{Contrastive Learning \and Multiple Instance Learning \and Weakly Supervised Learning \and Histopathological Images.}

\end{abstract}

\section{Introduction}
Deep learning has been widely utilized in medical image analysis, particularly for histopathological image classification tasks. Traditional deep learning-based image classification requires training on a large dataset in a fully supervised manner, where each image is assigned a label during training. However, this requirement poses significant challenges in the medical domain, as labeling a vast number of images demands substantial effort from experts. To address this issue, multiple instance learning (MIL) \cite{Shao2021-ug-transmil,Li2020-da-dsmil,zhang2022dtfd} methods have been proposed to enable weakly supervised training, thereby reducing the burden. 

In multiple instance learning (MIL), instead of assigning labels at the instance level, labels are provided at the bag level. Each bag contains multiple instances. A bag is labeled as positive if it contains at least one positive instance and is labeled as negative if all instances within it are negative. The MIL framework is particularly useful in histopathological image classification, where labels are assigned only at the whole slide image (WSI) level. Typically, each slide can be divided into thousands of patches, which serve as inputs for deep learning models. A slide is considered a positive slide if a portion of its patches are positive.

Due to computational constraints and the need to ensure the stability of MIL models under weak supervision, image patches are typically encoded into fixed embedded features before applying MIL methods. In attention-based MIL (AB-MIL) \cite{Ilse2018-qr-abmil}, an attention-based operator is introduced to assign different attention weights to patches within a bag. The final representation of a bag (slide) is obtained as a weighted sum of the feature embeddings from its instances (patches). The attention mechanism enables the model to identify key instances that contribute to the slide-level labels, thereby making a greater impact on the embedded features of slides during MIL training. 
Double-Tier Feature Distillation Multiple Instance Learning (DTFD-MIL) \cite{zhang2022dtfd} further enhances AB-MIL by introducing the concept of pseudo-bags to artificially increase the number of training bags. Each bag is split into a certain number of pseudo-bags, and instances are randomly assigned to these pseudo-bags. 

\begin{figure}
\centering
\includegraphics[width=0.6\textwidth]{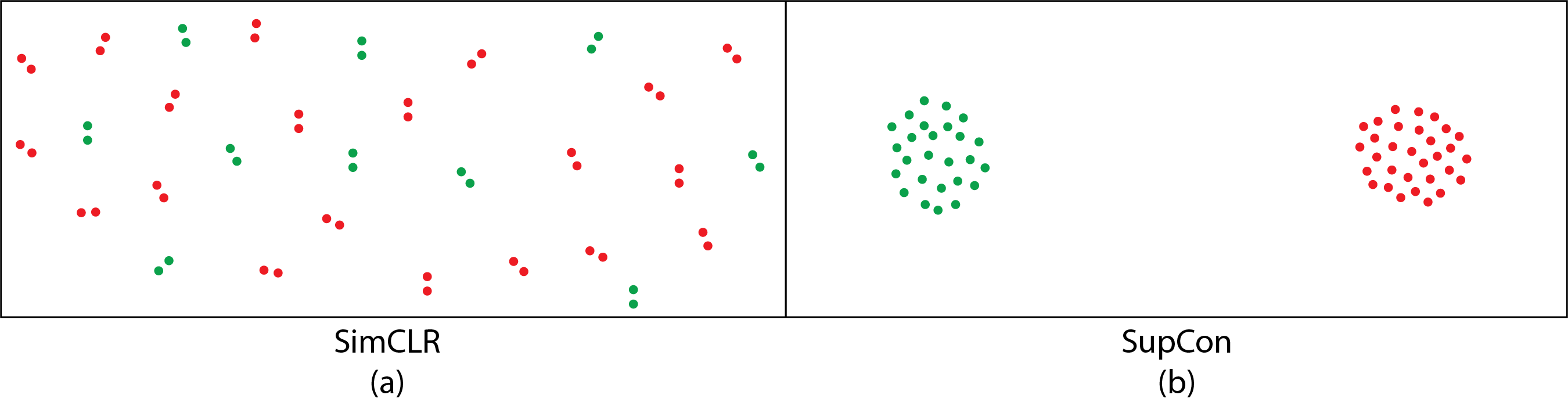}
\caption{Example of feature distributions after pre-training on (a) SimCLR and (b) SupCon. Different colors represent samples with different labels. } \label{fig_supcon}
\end{figure}

While the MIL structure directly influences MIL classification results, the quality of patch feature embeddings is crucial for the success of MIL classification. These features encapsulate the characteristics of instances in a way that facilitates learning the relationships between instances and the overall bag label. High-quality embedded features ensure that the aggregation process captures relevant patterns, thereby contributing effectively to bag-level classification. Although ImageNet pre-trained encoders are commonly used to generate patch features \cite{zhang2022dtfd}, the fact that ImageNet consists of natural images introduces a domain shift when these encoders are applied to histopathological images. An alternative way is to generate features by foundation models pre-trained on large histopathological images. \cite{Xu2024-kh-gigapath,Chen2024-es-uni2h} However, those foundation models are also pre-trained on histopathological datasets from other resources, which still introduces domain shifts to some extent. In this paper, we try to explore more effective approaches to pre-train encoders on local training datasets. Self-supervised contrastive learning \cite{Chen2021-gz-mocov3,Chen2020-ta-simclr2,zhang2022stain,manoochehrisra,zhang2025class} is a powerful method for pre-training encoders without the need for labels. Different views of the same sample are generated through image augmentation. Positive pairs are defined when two views originate from the same original sample, whereas negative pairs are formed when they come from different original samples. The contrastive loss function aims to minimize the distance between positive pairs while maximizing the distance between negative pairs. In SimCLR \cite{Chen2020-pf-simclr1}, given a batch of N samples, each sample is augmented twice, resulting in a total of 2N samples. The contrastive loss term between features from sample i and sample j is defined as:

\begin{equation}
\begin{aligned}
\ell_{i,j}=-\log{\frac{\exp(sim(z_{i},z_{j})/\tau)}{\sum_{k=1}^{2N}\mathbf{1}_{[k\not=i]}\exp(sim(z_{i},z_{k})/\tau)}}
\label{eq:SimCLR_term}
\end{aligned}
\end{equation}
where $sim(u,v)=u^{\top}v/\|u\|\|v\|$, $z_i$ and $z_j$ form a positive pair when they originate from the same original sample. An example of the distribution of learned features through self-supervised contrastive learning can be found in Fig.~\ref{fig_supcon} (a).

Supervised contrastive learning (SupCon) \cite{Khosla2020-yl-supcon} extends the self-supervised contrastive approach to a fully supervised setting by utilizing instance-level labels. The loss function is defined as:

\begin{equation}
\mathcal{L}^{sup} = \sum_{i \in I} \frac{-1}{|P(i)|} \sum_{p \in P(i)}{\log \frac{\exp(z_{i}\cdot z_{p}/\tau)}{\sum_{a \in A(i)}{\exp{(z_{i} \cdot z_{a} / \tau)}}}}
\label{eq:SuoConLoss}
\end{equation}
where $z_i$ and $z_p$ form a positive pair if they originate from samples that share the same label, $A(i)$ is the set of all samples except $i$, while $P(i)$ only contains samples that share the same label. $\tau$ is a temperature hyperparameter. SupCon has also been evaluated in transfer learning experiments on CIFAR-10, CIFAR-100, and ImageNet, where the encoder pre-trained with SupCon was frozen, and an additional classification layer was appended to perform classification training on the same dataset. Results \cite{Khosla2020-yl-supcon} have demonstrated that SupCon outperforms self-supervised contrastive learning and even traditional end-to-end training using cross-entropy loss \cite{Shannon1997-gr}. Fig.~\ref{fig_supcon} (b) demonstrates the ideal distribution of learned features after SupCon pre-training. Despite these promising results, the SupCon method that requires instance-level supervision still faces theoretical challenges in MIL settings, where only bag-level labels are available.
 
\begin{figure}
\centering
\includegraphics[width=0.9\textwidth]{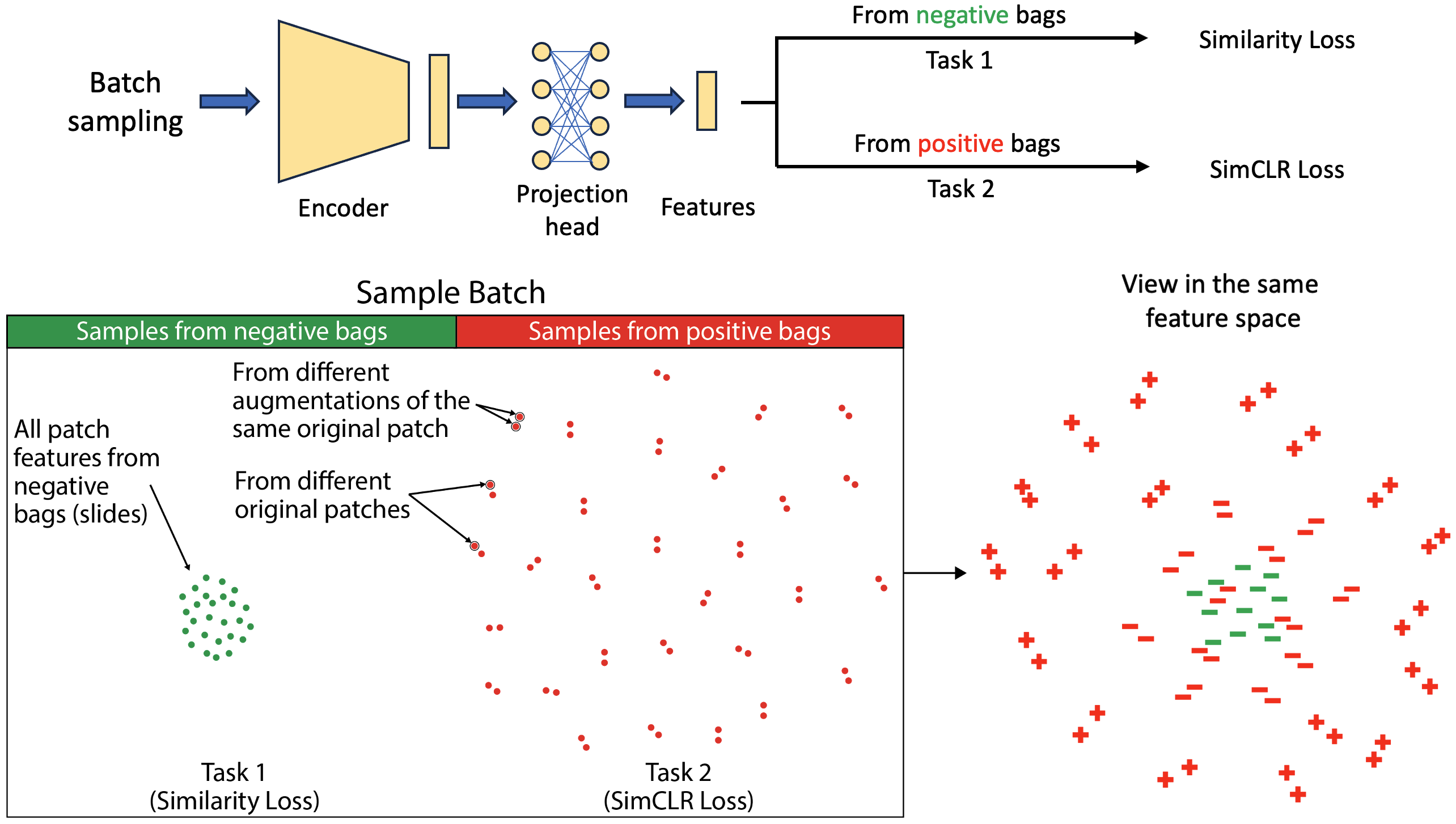}
\caption{Main idea of WeakSupCon. The samples in a batch are divided into distinct contrastive learning tasks based on their bag labels. Different colors denote different bag labels, while the symbols '+' and '-' indicate the actual instance labels. } \label{fig_weaksupcon}
\end{figure}

\section{Method}
Inspired by Supervised Contrastive Learning (SupCon), rather than pursuing a complex end-to-end multiple instance learning (MIL) solution, we propose weakly supervised contrastive learning (WeakSupCon) to enhance encoder pre-training by transitioning from a self-supervised to a weakly supervised approach. 

In our method, we treat all image patches within negative bags as negative samples based on the definition of negative bags. However, determining the true labels of patches within positive bags in pre-training is challenging, as they contain both positive and negative samples. To address this, we propose a multi-task learning approach that assigns different tasks to patch instances based on their bag labels. Fig.~\ref{fig_weaksupcon} illustrates the main concept of WeakSupCon. For each training batch during pre-training, we split samples into separate groups based on their bag labels. In the negative group, we apply a supervised version of the contrastive learning task to minimize the distances among all features from samples with negative bag labels. In detail, we introduce a Similarity Loss by taking only the positive pair component from the SupCon Loss \eqref{eq:SuoConLoss} to maximize the similarity among all the features from negative bags, as described below:
\begin{equation}
\mathcal{L}^{Similarity} = \sum_{i \in Neg} \frac{-1}{|Neg|} \sum_{j \in Neg, j \neq i}{z_{i}\cdot z_{j}/\tau}
\label{eq:SimilarityLoss}
\end{equation}
Where $Neg$ represents the negative group. In the positive group, we apply self-supervised contrastive learning (SimCLR) by defining the loss as follows:
\begin{equation}
\mathcal{L}^{SimCLR} = \sum_{i \in Pos} \ell_{i, p(i)}
\label{eq:SimCLRLoss}
\end{equation}
where $Pos$ represents the positive group, $p(i)$ denotes another feature generated from same original sample as $i$ but with a different augmentation, $\ell_{i, p(i)}$ is defined as in equation \eqref{eq:SimCLR_term}. Unlike SupCon, our method doesn't require that positive features cluster together. In total, our WeakSupCon loss will be:
\begin{equation}
\mathcal{L}^{WeakSupCon} = \mathcal{L}^{Similarity} + \mathcal{L}^{SimCLR}
\label{eq:WeakSupConLoss}
\end{equation}
The two groups are trained separately but simultaneously. Fig.~\ref{fig_weaksupcon} further illustrates the goal of WeakSupCon. When demonstrating the features from both groups in the same feature space, the negative samples within positive bags tend to move closer to the negative samples in negative bags, as Similarity Loss encourages the clustering of negative samples. In contrast, the positive samples in positive bags tend to move further away from negative samples due to the effect of the SimCLR loss. This separation makes MIL models more likely to locate or assign higher attention weights to positive samples.


\section{Experiments}
\subsection{Datasets}
We conducted experiments on three datasets, including Camelyon16 dataset \cite{camelyon16-yf-website,Ehteshami-Bejnordi2017-ai-camelyon16}, renal vein thrombosis (RVT) dataset, and kidney metastasis dataset. The Camelyon16 is a public dataset designed for slide-level classification between normal slides and metastatic cancer slides in lymph nodes, with 270 slides in the training set. The remaining two datasets are our institutional datasets that are available with data transfer agreements by contacting the authors.


In the renal vein thrombosis (RVT) dataset, case-level (patient-level) labels are provided instead of slide-level labels like in the Camelyon16 dataset, with each case usually containing multiple slides. The task is to predict the presence of RVT. The dataset consists of 74/12/18 negative cases and 31/8/11 positive cases in the training, validation, and test sets, respectively. In total, the training set contains 862 slides. We generated 1,518,872 patches with negative bag labels and 695,439 patches with positive bag labels in the training set for pre-training. 

The kidney metastasis dataset contains only case-level labels as well, with the goal of predicting metastasis. The training set consists of 15 negative cases and 42 positive cases. The validation set includes 10 negative cases and 26 positive cases. The test set contains 13 negative cases and 33 positive cases. In the training set, the negative cases provide 107 slides, from which 383,725 patches were generated. The positive cases provide 393 slides, with 930,939 generated patches. 

Among the datasets, the proportion of positive regions in positive slides varies significantly. In Camelyon16, it is estimated that less than 10\% of the foreground area in a positive slide contains cancerous regions \cite{Shao2021-ug-transmil,Li2020-da-dsmil}, whereas in the RVT dataset, positive regions occupy a much larger proportion in the positive slides. 

\subsection{Experiment settings}
We first pre-trained encoders by feature representation learning with various contrastive learning settings, including self-supervised learning (SimCLR, MoCo v3), supervised learning (SupCon), and weakly supervised learning (our WeakSupCon). For SupCon pre-training, all patch samples in positive bags were assigned positive pseudo-labels. Due to the limit of our computing resources, we used a batch size of 512. Compared to ResNet18 \cite{He2015-zb-resnet}, we found that using ViT-tiny \cite{dosovitskiy2020image} as the backbone was challenging across all contrastive learning settings. ResNet18 achieved around 10\% accuracy advantage in downstream MIL tasks in all settings. Consequently, we adopted ResNet18 as the backbone for all models.


After encoder pre-training, we extracted features for all patches in the training, validation, and test sets and evaluated different encoder models using DTFD-MIL, considering its high performance and popularity. 
We set the number of pseudo-bags to 5 for the Camelyon16 dataset and to 30 for the remaining two datasets to leverage the large number of positive patches in each bag. For comparison, we also generated features by pathology foundation models pre-trained on extremely large datasets, including Prov-GigaPath \cite{Xu2024-kh-gigapath} and UNI2-h \cite{Chen2024-es-uni2h}. 

Our experiments were conducted on NVIDIA RTX A6000 GPUs. Each epoch requires approximately 35 GB of GPU memory and 2 hours. The memory and running time of WeakSupCon are similar to SimCLR and SupCon. We repeated each MIL experiment three times to get balanced accuracy, accuracy, and AUC. Our code is available at \url{github.com/BzhangURU/Paper_WeakSupCon}

\begin{table}[t]
\centering
\caption{Comparison between our WeakSupCon model and other contrastive learning approaches, including self-supervised contrastive learning models (MoCo v3, SimCLR) and supervised contrastive learning model (SupCon). 
}\label{tab1_contrastive_learning}
\begin{tabular}{|l|l|l|l|}
\hline
\textbf{Encoder} &  \textbf{Balanced acc} & \textbf{Accuracy} & \textbf{AUC}\\
\hline
\multicolumn{4}{|c|}{MIL Results on Camelyon16 dataset} \\ 
\hline
MoCo v3  &  $0.9051 \pm 0.0200 $ & $0.9199 \pm 0.0195$ & $0.9238 \pm 0.0050$\\
\hline
SimCLR  &  $0.8928 \pm 0.0050 $ & $0.9095 \pm 0.0089$ & $0.9130 \pm 0.0048$\\
\hline
SupCon  &  $0.8760 \pm 0.0191 $ & $0.9018 \pm 0.0161$ & $0.8792 \pm 0.0077$\\
\hline
WeakSupCon  &  {\bfseries 0.9265}$\pm 0.0036 $ & {\bfseries 0.9431}$\pm 0.0044$ & {\bfseries 0.9694}$\pm 0.0018$\\
\hline
\multicolumn{4}{|c|}{MIL Results on renal vein thrombosis (RVT) dataset} \\ 
\hline
MoCo v3  &  $0.7012 \pm 0.0466 $ & $0.7241 \pm 0.0345$ & $0.7610 \pm 0.0556$\\
\hline
SimCLR  &  $0.6785 \pm 0.0239 $ & $0.7012 \pm 0.0526$ & $0.7592 \pm 0.0466$\\
\hline
SupCon  &  $0.7281 \pm 0.0368 $ & $0.7356 \pm 0.0527$ & $0.8266 \pm 0.0304$\\
\hline
WeakSupCon  &  {\bfseries 0.8014}$\pm 0.0130 $ & {\bfseries 0.8046}$\pm 0.0199$ & {\bfseries 0.8771}$\pm 0.0177$\\
\hline
\multicolumn{4}{|c|}{MIL Results on kidney metastasis dataset} \\ 
\hline
MoCo v3  &  $0.8633 \pm 0.0135 $ & $0.8261 \pm 0.0000$ & $0.8982 \pm 0.0048$\\
\hline
SimCLR  &  $0.8939 \pm 0.0000 $ & $0.8478 \pm 0.0000$ & $0.9192 \pm 0.0088$\\
\hline
SupCon  &  $0.8858 \pm 0.0000 $ & {\bfseries 0.8696}$\pm 0.0000$ & $0.9176 \pm 0.0016$\\
\hline
WeakSupCon  &  {\bfseries 0.9091}$\pm 0.0000 $ & {\bfseries 0.8696}$\pm 0.0000$ & {\bfseries 0.9277}$\pm 0.0016$\\
\hline

\end{tabular}
\end{table}

\subsection{Experiment results}
Table~\ref{tab1_contrastive_learning} presents the results from different encoder pre-training models across the three datasets. It is worth noting that the SupCon and our WeakSupCon are based on SimCLR, with modifications to the contrastive loss functions. As a result, it provides a fair comparison among different contrastive loss designs in MIL tasks. Overall, our WeakSupCon method achieves the best performance across all three datasets. Interestingly, SupCon performs worse than SimCLR on the Camelyon16 dataset but outperforms SimCLR on the RVT dataset. This discrepancy can be attributed to the proportion of positive patches within positive slides. In Camelyon16, fewer than 10\% of patches in tumor slides contain actual tumor regions, leading to poor pseudo-labeling in SupCon. In contrast, WeakSupCon does not assign pseudo-labels to patches in tumor slides. Instead, it allows the loss functions to automatically cluster negative samples while enabling positive samples to generate more distinct features. In the RVT dataset, where a larger percentage of patches contain positive regions, the pseudo-labeling in SupCon is more accurate. The superior performance of SupCon in the RVT dataset also underscores the potential of contrastive learning in a supervised setting compared to its self-supervised counterpart. On the kidney metastasis dataset, we observed that the accuracies are the same across three runs for all contrastive learning models, probably due to the small number of cases on the test set and the robust feature representations produced by contrastive learning.

\begin{figure}[h]
\centering
\includegraphics[width=0.9\textwidth]{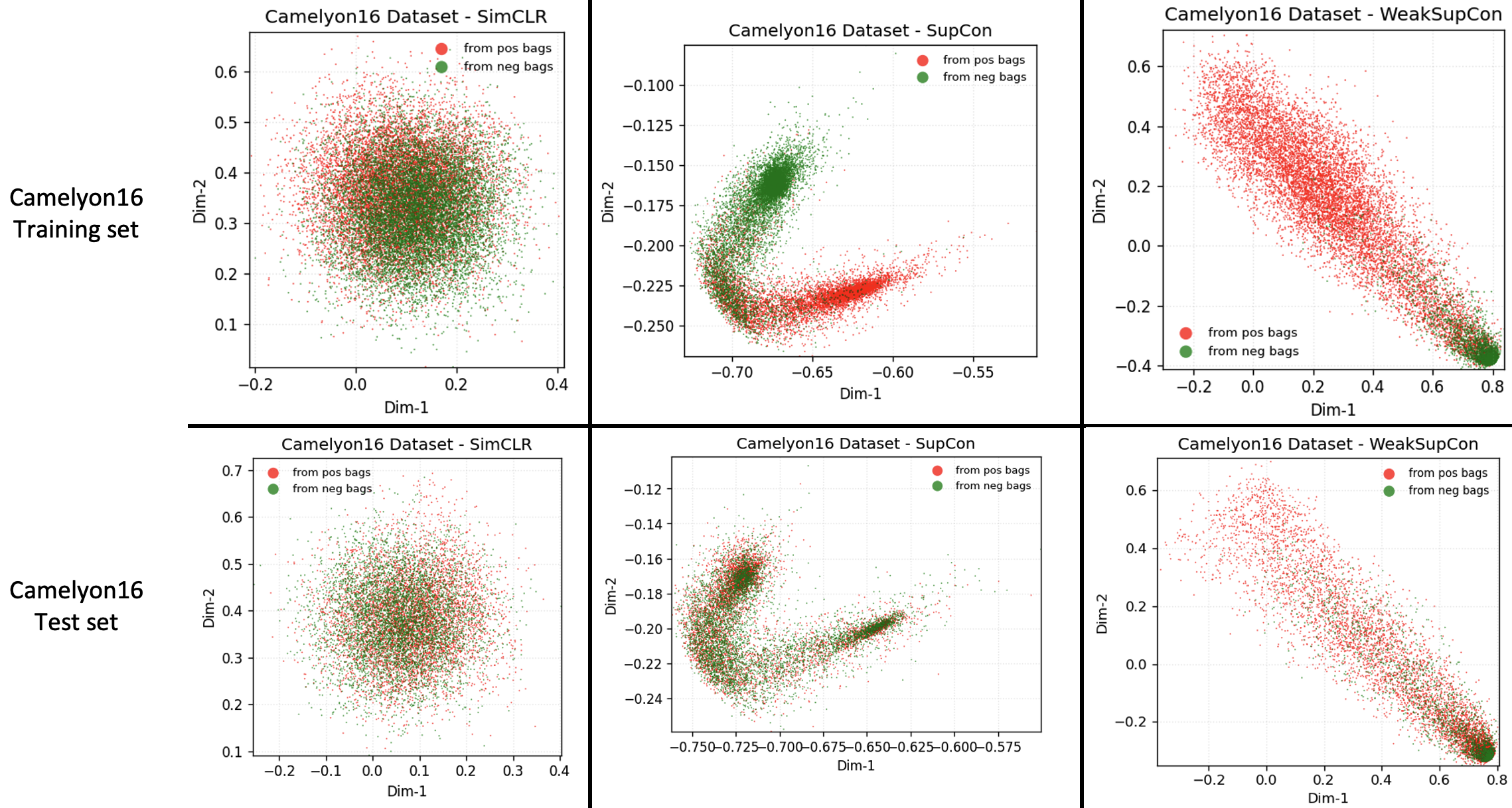}
\caption{PCA visualization of feature distributions on the Camelyon16 dataset after pre-training on the training set. The features were downsampled from the whole training set or test set for better visualization.} \label{fig_PCA_only_Camelyon16}
\end{figure}

To better visualize the learned features from different models after encoder pre-training, we applied principal component analysis (PCA) to reduce the features to two dimensions on the Camelyon16 dataset. The feature distributions after pre-training by SimCLR, SupCon, and WeakSupCon can be found in Fig.~\ref{fig_PCA_only_Camelyon16}. Based on these figures, even though the features pre-trained by a self-supervised learning method like SimCLR demonstrate large variations, indicating an abundant amount of information, the distribution differences between features from negative and positive bags are not obvious, which introduces challenges in downstream classification. The features pre-trained by SupCon show distinct distributions between negative bags and positive bags in the training set. However, the unavoidable errors caused by incorrect pseudo-labels lead to overfitting and affect generalization, especially for datasets with only a small portion of positive instances in positive bags. Consequently, as shown in the figures, SupCon fails to produce distinct feature distributions between negative and positive bags in the test set of the Camelyon16, and the features are totally mixed. Moreover, the variations of these features are much smaller compared to those from SimCLR, only varying between -0.75 and -0.5 in the first dimension in the Camelyon16 training set, as illustrated in Fig.~\ref{fig_PCA_only_Camelyon16}. In contrast, our WeakSupCon first generates distinct feature distributions between negative bags and positive bags by assigning different loss functions according to the bag labels. The Similarity loss encourages instance features with negative labels to cluster, while the SimCLR loss encourages features with positive labels to exhibit greater variation, enabling these features to receive higher attention in downstream MIL tasks. The variations of features are even greater than those obtained by SimCLR, with values ranging between -0.3 and 0.8 in the first dimension in the Camelyon16 dataset. In WeakSupCon, the features from positive bags and negative bags also appear more mixed in the Camelyon16 test set than in the training set. However, this does not necessarily indicate that the model is not generalizing well. In fact, all patches from the test set are new to the model, and more than 90\% of patches from positive bags are actually negative. As a result, those patches will theoretically mix with negative patches from negative bags. In theory, if a sample from positive bag (a red dot) is closer to green dots, it is more likely that the sample is also negative. What we only expect to see is that a small portion of patches from positive bags do not mix with patches from negative bags, and those patches are likely the true positive patches in the positive bags. 

\begin{table}[t]
\centering
\caption{Comparison between our WeakSupCon model and foundation models, including state-of-the-art pathology foundation models (Prov-GigaPath and UNI2-h). 
}\label{tab2_foundation_models}
\begin{tabular}{|l|l|l|l|}
\hline
\textbf{Encoder} &  \textbf{Balanced acc} & \textbf{Accuracy} & \textbf{AUC}\\
\hline
\multicolumn{4}{|c|}{MIL Results on Camelyon16 dataset} \\ 
\hline
ImageNet pre-trained & $0.8034 \pm 0.0211$ & $0.8346 \pm 0.0119$ & $0.8530 \pm 0.0186$ \\
\hline
Prov-GigaPath  &  {\bfseries 0.9469}$\pm 0.0036 $ & {\bfseries 0.9586}$\pm 0.0044$ & $0.9757 \pm 0.0009$\\
\hline
UNI2-h  &  $0.9461 \pm 0.0032 $ & $0.9561 \pm 0.0044$ & {\bfseries 0.9782}$\pm 0.0016$\\
\hhline{|=|=|=|=|}
WeakSupCon  &  $0.9265 \pm 0.0036 $ & $0.9431 \pm 0.0044$ & $0.9694 \pm 0.0018$\\
\hline
\multicolumn{4}{|c|}{MIL Results on renal vein thrombosis (RVT) dataset} \\ 
\hline
ImageNet pre-trained &  $0.6330 \pm 0.0238$ & $0.6322 \pm 0.0199$ & $0.6280 \pm 0.0480$ \\
\hline
Prov-GigaPath &  $0.7096 \pm 0.0140$ & $0.7126 \pm 0.0199$ & $0.7576 \pm 0.0101$ \\
\hline
UNI2-h &  $0.6524 \pm 0.0229$ & $0.5977 \pm 0.0199$ & $0.6229 \pm 0.0191$ \\
\hhline{|=|=|=|=|} 
WeakSupCon  &  {\bfseries 0.8014}$\pm 0.0130 $ & {\bfseries 0.8046}$\pm 0.0199$ & {\bfseries 0.8771}$\pm 0.0177$\\
\hline
\multicolumn{4}{|c|}{MIL Results on kidney metastasis dataset} \\ 
\hline
ImageNet pre-trained  &  $0.7848 \pm 0.0291 $ & $0.7246 \pm 0.0251$ & $0.8415 \pm 0.0142$\\
\hline
Prov-GigaPath  &  $0.8862 \pm 0.0076 $ & $0.8478 \pm 0.0218$ & $0.9192 \pm 0.0178$\\
\hline
UNI2-h  &  $0.8889 \pm 0.0175 $ & $0.8406 \pm 0.0251$ & $0.9161 \pm 0.0185$\\
\hhline{|=|=|=|=|}
WeakSupCon  &  {\bfseries 0.9091}$\pm 0.0000 $ & {\bfseries 0.8696}$\pm 0.0000$ & {\bfseries 0.9277}$\pm 0.0016$\\
\hline

\end{tabular}
\end{table}

We also compared our WeakSupCon with foundation models. As shown in Table~\ref{tab2_foundation_models}, WeakSupCon always significantly outperforms the ImageNet pre-trained backbone, highlighting the importance of feature encoders for downstream MIL tasks. Additionally, two state-of-the-art pathology foundation models were included in the experiments. We found that WeakSupCon pre-trained on just one GPU outperforms Prov-GigaPath and UNI2-h in most cases, despite that they were pre-trained on more than 1.3 billion or 200 million pathology patches using much more powerful computing resources and substantially larger backbones.

\section{Conclusion}
In this paper, we propose Weakly Supervised Contrastive Learning (WeakSupCon) for encoder pre-training by leveraging bag-level labels. The features learned through WeakSupCon further enhance downstream MIL performance compared to self-supervised contrastive learning, highlighting the potential for more accurate medical diagnosis under limited training labels and computing resources.

    

\begin{credits}
\subsubsection{\ackname} The experiments were funded by NIH/NCI 1R21CA277381, DoD HT94252410186, and Department of Veterans Affairs I01CS002622. We also acknowledge the support of the Computational Oncology Research Initiative (CORI) at the Huntsman Cancer Institute, ARUP Laboratories, and the Department of Pathology at the University of Utah.

\subsubsection{\discintname}
The authors have no competing interests to declare that are
relevant to the content of this article.
\end{credits}

%
%
%
\bibliographystyle{splncs04}
\bibliography{refs}


\end{document}